# DeepNose: An Equivariant Convolutional Neural Network Predictive Of Human Olfactory Percepts

Sergey Shuvaev
*Department of Neuroscience*
*Cold Spring Harbor Laboratory*
Cold Spring Harbor, USA
sshuvaev@cshl.edu

Khue Tran
*Department of Neuroscience*
*Cold Spring Harbor Laboratory*
Cold Spring Harbor, USA
ktran@cshl.edu

Khristina Samoilova
*Department of Neuroscience*
*Cold Spring Harbor Laboratory*
Cold Spring Harbor, USA
samoilo@cshl.edu

Cyrille Mascart
*Department of Neuroscience*
*Cold Spring Harbor Laboratory*
Cold Spring Harbor, USA
mascart@cshl.edu

Alexei Koulakov
*Department of Neuroscience*
*Cold Spring Harbor Laboratory*
Cold Spring Harbor, USA
koulakov@cshl.edu

*Abstract*— The olfactory system employs responses of an ensemble of odorant receptors (ORs) to sense molecules and to generate olfactory percepts. Here we hypothesized that ORs can be viewed as 3D spatial filters that extract molecular features relevant to the olfactory system, similarly to the spatio-temporal filters found in other sensory modalities. To build these filters, we trained a convolutional neural network (CNN) to predict human olfactory percepts obtained from several semantic datasets. Our neural network, the DeepNose, produced responses that are approximately invariant to the molecules' orientation, due to its equivariant architecture. Our network offers high-fidelity perceptual predictions for different olfactory datasets. In addition, our approach allows us to identify molecular features that contribute to specific perceptual descriptors. Because the DeepNose network is designed to be aligned with the biological system, our approach predicts distinct perceptual qualities for different stereoisomers. The architecture of the DeepNose relying on the processing of several molecules at the same time permits inferring the perceptual quality of odor mixtures. We propose that the DeepNose network can use 3D molecular shapes to generate high-quality predictions for human olfactory percepts and help identify molecular features responsible for odor quality.

## I. INTRODUCTION

The olfactory system relies on hundreds of odorant receptor (OR) types to recognize millions of volatile molecules [1]. ORs are the specialized proteins that can be activated by odorant molecules (Fig. 1). The OR family includes ~1,000 receptors in mice and rats, ~50 receptors in the fruit fly, and ~350 functional receptors in humans [2, 3]. ORs exhibit broad and specific combinatorial activation by odorants, leading to the ability of the olfactory system to recognize a wide range of stimuli [3-6]. ORs are expressed by the olfactory sensory neurons (OSNs) in the olfactory epithelium (OE). Each mature OSN expresses one and only one functional OR type (Fig. 1). ORs transduce odor binding to OSN electrical activity, which represents the odorant to the olfactory networks.

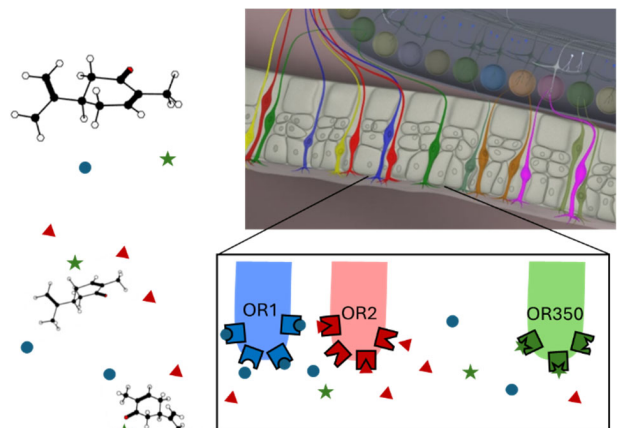

Fig. 1. Molecules are sensed by specialized proteins, called odorant receptors (ORs), that are embedded in the membranes of neurons in the olfactory epithelium called Olfactory Sensory Neurons (OSNs, colored cells). Each OSN makes only one type of OR, defined by the genetic sequence, out of ~350 possibilities in humans. ORs can recognize shapes and distinct chemical groups in the odorant molecules, become activated, and convey the levels of activation to the responses of OSNs. OSNs make synaptic connections to the olfactory networks in the brain which can recognize odorants.

Here, we train deep convolutional neural networks (CNNs) [7] to represent the space of molecules and use the derived representations to predict human perceptual responses. CNNs use ensembles of spatial filters of increasing complexity to extract features relevant to representing the input objects. Networks based on CNNs are the best-performing method for image recognition, many of which perform better at classifying images than humans [8, 9]. The performance of CNNs in 2D image recognition tasks suggests that they are well-suited for learning other types of spatial data, such as molecular structures. Computational chemistry has used a diverse set of approaches to represent molecules or ligands in neural

NIH BRAIN Initiative grant U19NS112953.

networks. Molecules have been described by SMILES strings [10], 2D images, graphs, and 3D grids [11-14]. In olfaction, previous studies have represented odorant molecules using molecular fingerprints, SMILES strings, or graphs [15-17]. In this study, we extend our previous work on the CNNs that can recognize molecules and predict olfactory percepts [18] to four datasets (Fig. 2) [19-21].

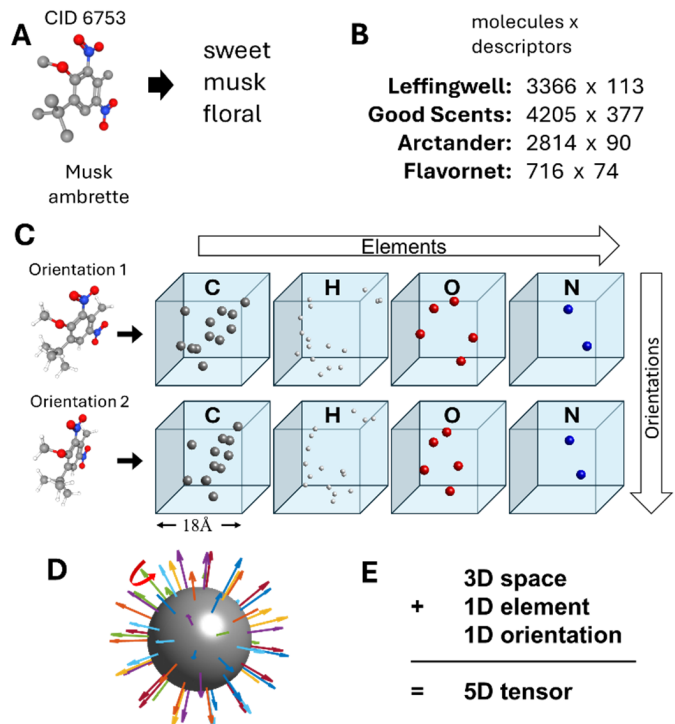

Fig. 2. Our study is based on data from four datasets containing semantic descriptors for thousands of molecules. (A) An example dataset entry for one molecule. The 3D structure has been obtained from PubChem. (B) Four datasets used in our study. The first and second numbers for each dataset define the number of molecules included and the number of semantic descriptors, respectively. (C) Our model represents each molecule as an array of 3D images. For each molecule, the images differ in the chemical element present at each point and the molecule's orientation. (D) Orientations are sampled from the 64 angles uniformly covering the 3D sphere and 10 axial rotations around each of these angles (red arrow), 640 orientations overall. (E) Each of our molecules is represented by a 5D tensor. Three dimensions included 3D spatial dimensions, one dimension represented atomic features and the last dimension represented 640 molecules' orientations. Due to the dense sampling of the molecules' orientation, responses of our network were approximately equivariant.

In designing our neural network, our main hypothesis is that the OR ensemble functions as a set of 3D spatial filters, interacting with molecular structures in the 3D space. In this framework, molecules act as 3D images, while ORs serve as analogs of receptive fields, such as edge detectors in the visual system [22]. Both visual receptive fields and ORs have been shaped by evolution to detect relevant features present in their stimuli. We propose that an efficient OR ensemble can be found in a computational model using standard machine learning techniques, such as backpropagation [23], if we optimize it to accurately represent the molecular space. This approach enables machine learning to uncover factors influencing OR evolution, build models for OR-ligand binding, and derive human olfactory percepts. Overall, our goal is to implement a neural network that approximates the architectural features of the olfactory system.

## II. RESULTS

Much of what is known about the human sense of smell comes from semantic datasets. These datasets typically associate molecules with semantic descriptors – words – that define the perceptions evoked by smelling them (Fig. 2A). The descriptors are often binary, indicating the applicability of each word (e.g. "sweet") without quantifying its intensity. Several such datasets, containing thousands of molecules, include Leffingwell [21], Good Scents [20], Arctander [19], and Flavornet [24]. To link the percepts to molecular structures of odorants, we used PubChem, a publicly available resource offering 3D conformations and additional properties of molecules [25]. For our neural network, we used the 3D structure of the odorant molecule as an input and the set of semantic descriptors defining corresponding percepts as output.

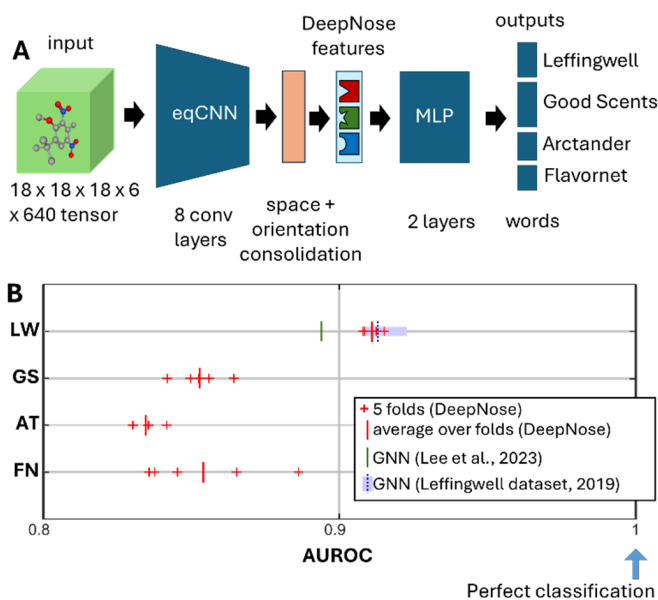

Fig. 3. The architecture of the DeepNose network and the results of its training. (A) The network includes 8 convolutional layers (eqCNN), a space-orientation consolidation layer, and 2 fully connected layers (MLP). The network receives a 5D tensor representing each molecule. Every semantic descriptor in each dataset is treated as a separate output channel. The network structure is described in detail in Methods. (B) The results are described as the Area Under Receiver Operating Characteristic (AUROC) averaged for each dataset (LW-Leffingwell, GS-Good Scents, AT-Arctander, FN-Flavornet). Perfect/random predictions correspond to AUROC of 1 and 0.5 respectively. The results are obtained for held-out data separated into 5 non-overlapping folds. The results for each fold, the average over folds, and the previous SOTA results (Refs. [16, 21] and [26]) are shown by red crosses, red vertical lines, and green and blue lines respectively. The shaded blue region shows the statistical uncertainty reported in Ref. [26].

One of the key features of olfactory stimuli is that the perception of molecules is invariant with respect to their orientation. If, for example, the principal axis of a molecule happens to be horizontal in our 3D data, the output of the network has to be the same as if it were vertical. In other words, the output of the network has to be orientation invariant, i.e. independent of the input orientation. This property is usually attained using graph neural networks (GNNs), which receive the graphs of the molecules' connectivity defined by chemical bonds as the input.

That two atoms in the molecule form a bond is not dependent on the molecule's position or orientation. For this reason, connectivity is not sensitive to orientation and the outputs of GNNs inherit this property.

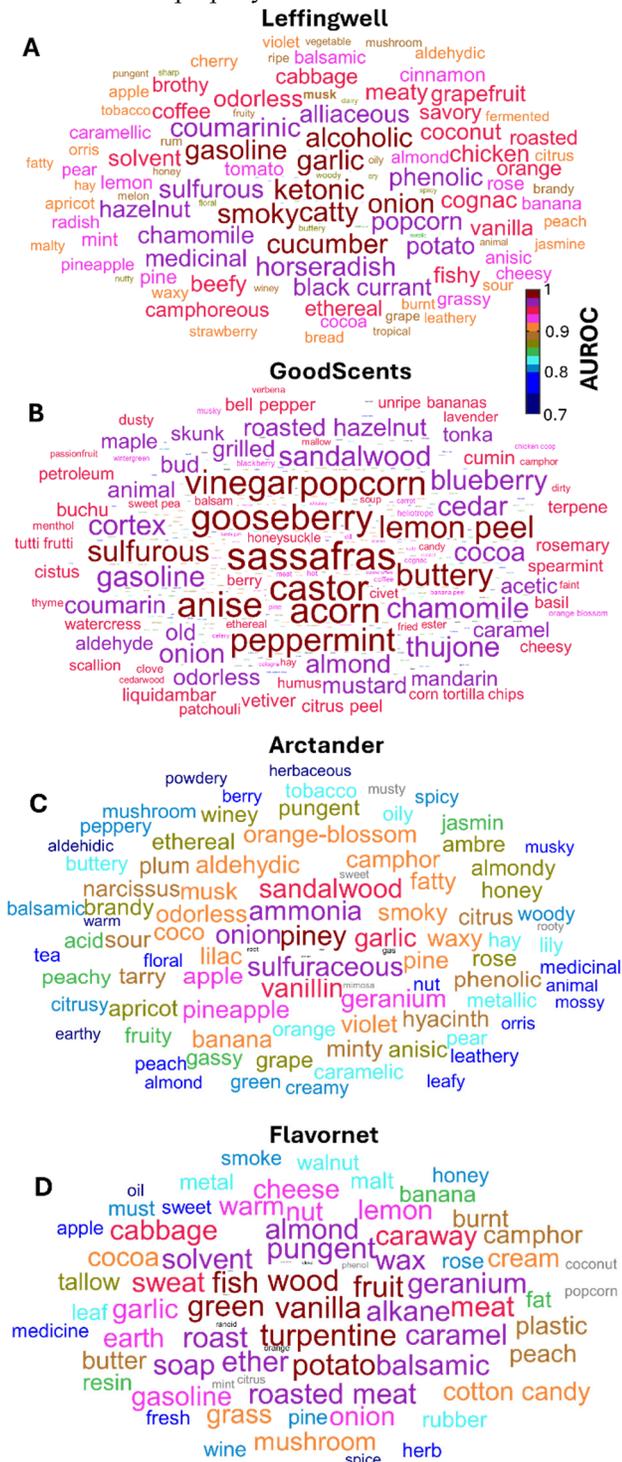

Fig. 4. Word clouds containing AUROC for individual semantic descriptors for all the four datasets. Individual AUROCs are color-coded as indicated and represented by the words' height.

Our goal was to design a network that reflects the architectural features of the olfactory system. Since it is unlikely that ORs have access to the information about a molecule's bond-based connectivity matrix, we had to find a different solution. Instead, we took advantage of the fundamental property of olfactory stimuli, that the same molecule is presented to the OR ensemble in multiple copies and can approach the ORs at multiple angles. This consideration determined the structure of the inputs into the neural network.

In our model, each molecule was presented to the network as a 5D tensor. The first three dimensions of this tensor are the spatial dimensions within which the molecular structure is located. These three dimensions are represented by a 3D raster which contains voxelated density distribution of the corresponding atoms (Fig. 2C). We used rasters that are 18Å×18Å×18Å in size with a voxel size of 1Å. The fourth dimension of the molecule's representation encodes the identity of the chemical element: C, H, O, N, S, Cl. This dimension is analogous to the RGB color space for images. The fifth dimension represents the molecule's orientation. We sampled the orientations of each molecule from three polar angles. 64 values of azimuth and elevation defined the position of the molecule's principal axis and covering the sphere (Fig. 2D). In addition, each axis direction was represented by 10 molecule's rotations around this direction. Overall, each molecule was presented to the network in 640 copies ($64 \times 10$) which differed in the molecule's orientation.

The 5D tensor representing the spatial structure of each molecule served as an input to the network. The goal of the deep neural network was to compute the multi-hot binary semantic perceptual descriptors for each molecule. Our feedforward neural net, which we called the DeepNose, contained two major blocks of layers. First, the spatial structures of odorants were put through a CNN which included 8 trainable 3D convolutional layers (Fig. 3A). The CNN operated on each orientation of the molecule independently of other orientations, as if it were a different molecule. The same set of convolutional filters was applied to all orientations allowing for weight sharing along this dimension. At the end of the convolutional block, the molecule was represented by 96 features which were defined for each voxel of a $2 \times 2 \times 2$ spatial cube and for each of the 640 orientations. At this point, we performed the averaging over the spatial and orientation dimensions to obtain the representation of molecules in terms of the 96 DeepNose features. We called this operation space and orientation consolidation (Fig. 3A). The resulting DeepNose features contained the properties of molecules that were learned to be predictive of the human olfactory percepts and therefore relevant to the olfactory system. The representation in terms of the DeepNose features was approximately invariant to the molecule's spatial shifts and rotations. Indeed, if the input molecule were rotated between two angles in our orientation grid, its representation just shifted on the grid producing no change in the average over the orientations. Our representation was therefore equivariant to discrete rotations and shifts. Consequently, the DeepNose features were invariant to these transformations [18]. To reflect this invariance property, we called our CNN block an equivariant CNN or eqCNN (Fig. 3A). After the consolidation layer, the DeepNose features represented the activations of the

3D spatial filters that were invariant to molecules' orientations and were predictive of the human olfactory percepts. We, therefore, argue that the activations of the 96 DeepNose features play a similar role to the responses of 350 human OR channels to odorants. The DeepNose features were then sent through two fully connected layers to produce the multi-hot binary semantic descriptors for each of the four datasets as separate channels. The fully connected part of the network was assumed to provide a model of the neural networks that receive inputs from the ORs [27].

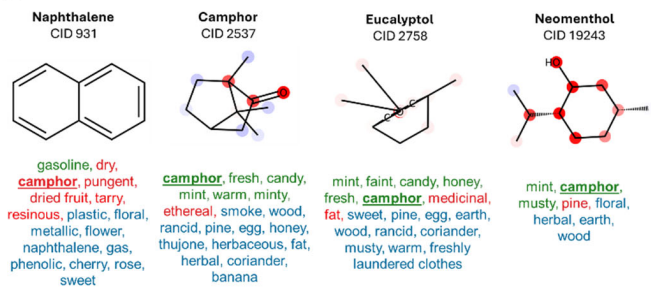

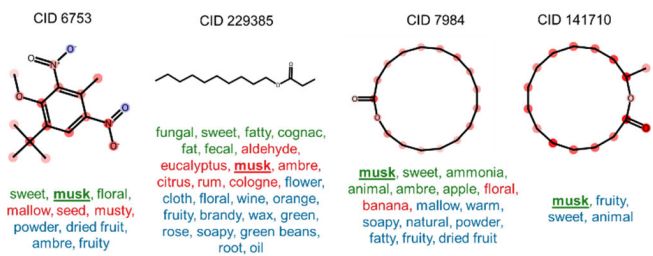

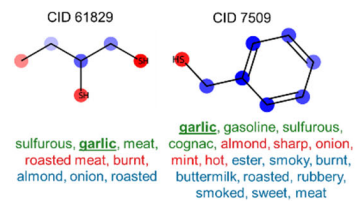 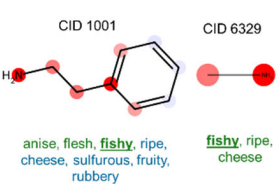

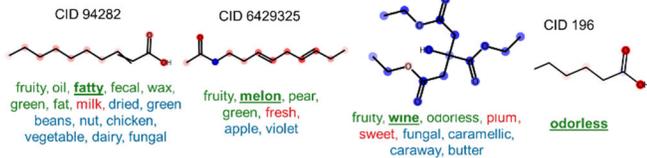

Fig. 5. Examples of the semantic descriptors predicted by our network and perceived by humans. The true positive/false negative/false positive predictions are shown by green/red/blue semantic descriptors for each molecule. This implies that green descriptors were correctly predicted, red descriptors were missed, and blue descriptors were additionally discovered by the DeepNose. (A) Molecules described as "camphor" (bold underlined word). Positive/negative contributions to the "camphor" descriptor are indicated by red/blue markers surrounding each atom. (B) Musks. The contributions to the "musk" descriptor (bold underlined) are indicated by color. (C) Other molecules. Contributions to each descriptor ("fatty" to "odorless") are shown by color for each atom. In many cases, such as the "naphthalene" descriptor in (A) and the "dairy" descriptor in (E, CID 94282) the additionally discovered predictions (blue) are meaningful.

After training, our model produced high-quality predictions for the human olfactory percepts. To describe the prediction accuracy, we used the average value of Area Under Receiver Operating Characteristic (AUROC), a measure frequently used to quantify predictions of sparse binary variables [16, 17, 21, 28]. AUROC values equal to 0.5/1.0 correspond to the random/perfect prediction. We evaluated AUROC for each dataset separately using held-out testing data. Specifically, we trained the DeepNose on 80% of the available molecules and tested its performance on 20% of the remaining molecules. To avoid potential testing biases, in evaluating our model, we used 5-fold cross-validation. These folds corresponded to 5 different splits of the set of molecules into training and testing data (80% and 20%) in such a way that the testing data from different folds did not overlap. This implies that each molecule belonged to a testing set for one of the folds. The AUROC values for different folds and the average over the testing data are shown in Fig. 3B and Table 1. For the four datasets in our study, the values of AUROC averaged over folds ranged between 0.835 and 0.911. For the best-performing dataset (Leffingwell), the performance of our network was within the range of the variance of the previously released State-of-the-Art (SOTA) result (Fig. 3B, blue dotted line) [26]. The SOTA result was obtained for a single 20%-fold of testing data. Overall, the performance of our network matches the SOTA results within the range of statistical significance. We also computed AUROC for every descriptor in each dataset (Fig. 4).

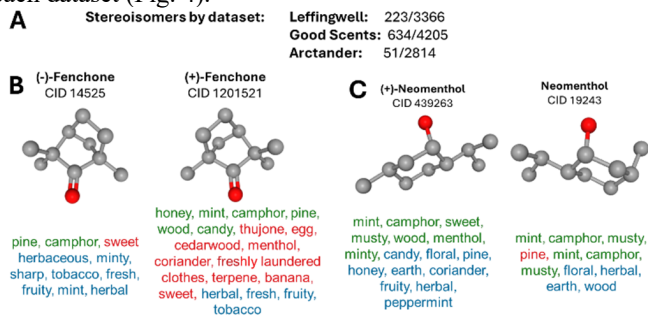

Fig. 6. CNN kernels in our model can distinguish stereoisomers. (A) Datasets and isomers. (B) DeepNose distinguishes enantiomers. (C) Stereoisomers.

*A. Contributions of molecular features to individual descriptors.*

What molecular elements lead to a particular percept? To address this question, we used the occlusion method [29, 30]. In this approach, individual atoms in a molecule were removed, the molecular structure was put through the network and the change in the network output was monitored. A large reduction in the output for some semantic descriptors produced by an atom removal suggested that this atom contributes strongly to the given descriptor. Examples of molecules with contributions of individual atoms are shown in Fig. 5.

TABLE I.   AUROC (AVERAGE OVER FOLDS)

| Dataset | GNN (SOTA) [26] | Average AUROC |
|---|---|---|
| Leffingwell [21] | 0.913 (0.908-0.923) | 0.911 |
| GoodScents [20] | | 0.853 |
| Arctander [19] | | 0.835 |
| Flavornet [24] | | 0.854 |

*B. Responses of DeepNose to stereoisomers.*

The DeepNose network can distinguish between the perceptions of stereoisomers. Since the DeepNose responds to the geometric shape of molecules in 3D space, it produces distinct responses to both enantiomers and other types of stereoisomers. This capability sets our model apart from naive GNNs and stems from its design, which mirrors its biological prototype. In the four datasets analyzed (Fig. 6A), we identified numerous stereoisomers that evoke different percepts. DeepNose consistently provided distinct outputs for pairs of isomers, including enantiomers (Fig. 6B,C). This demonstrates that DeepNose can approximate human responses to stereoisomers without requiring additional parameters beyond the molecules' 3D shapes.

*C. Application of the DeepNose network to mixtures.*

The architecture of the DeepNose network enables it to handle odorant mixtures containing multiple monomolecular components. As mentioned earlier, our equivarant CNN (eqCNN) processes each molecular orientation independently of others, treating each orientation as a distinct molecule. This characteristic allows different monomolecular components of a mixture to be passed through separate orientation channels. After averaging over space and orientations in the consolidation layer, we derive DeepNose features that describe the entire mixture.

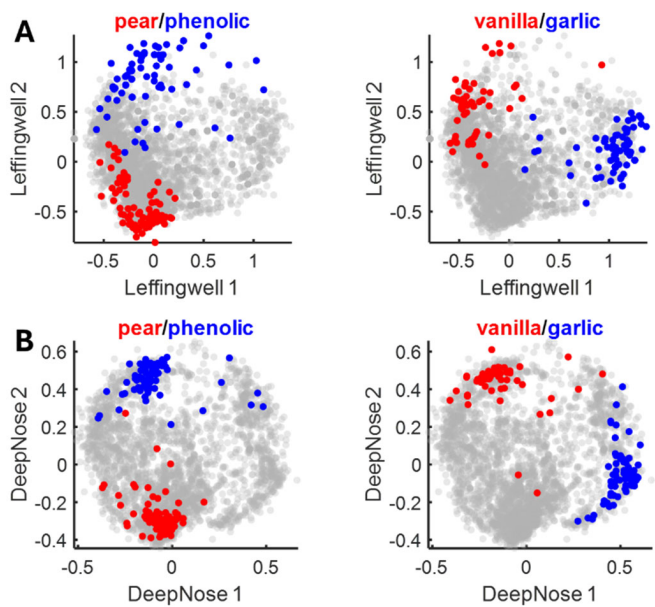

Fig. 7. Odor spaces defined for the Leffingwell dataset. (A) Embedding of 3366 odorants based on their semantic descriptors using the Isomap algorithm [31]. Odorants carrying semantic descriptors are colored as described in the title of each panel. (B) Embedding of the same set of odorants based on their DeepNose features. The features were embedded into 3D using Isomap and rotated to match the dimensions in (A). The layout of semantic descriptors is similar to (A).

From a practical point of view, these steps can be simplified. Since each orientation channel in the eqCNN operates independently and the averaging operation in the consolidation layer is linear, the DeepNose features for individual mixture components can be computed separately and then averaged to represent the entire mixture. This approach reduces the variance in the mixture representation caused by the limited number of orientation channels per molecule. Using this procedure, we analyzed four datasets containing perceptual distances between pairs of mixtures consisting of 1–43 monomolecular components [32-34]. For each mixture, we computed a vector of 96 DeepNose features and approximated perceptual dissimilarities between mixtures by Euclidean distances between these feature vectors. Without any training on mixture data, our model demonstrated a significant correlation (R=0.40) between the observed mixture dissimilarities and the distances between the DeepNose features (Fig. 8B). This result indicates that our neural network can effectively represent mixtures of monomolecular components using only their 3D molecular structures. Notably, the network parameters were obtained solely by training on monomolecular perceptual datasets. We propose that the DeepNose exhibits zero-shot generalizability across different data types, enabling it to transition from monomolecular odorants to mixtures.

*D. The significance of the DeepNose features*

What properties of odorants do the DeepNose features represent? To address this question, we constructed an odor space by computing a low-dimensional embedding of odorants. Odor spaces are the actively studied constructs [16, 18, 35-45] representing olfactory sensory objects within a common set of dimensions, analogous to the color space in the visual system.

First, we built the odor space based on human semantic descriptors using the Isomap algorithm [18, 31]. Briefly, pairs of molecules in the dataset (Leffingwell) were connected if they were described by the same word. The target distance between two connected molecules was decreased if they shared more semantic descriptors [18]. The Isomap algorithm discarded long links between molecules as unreliable and computed the geodesic distances between pairs of smells based on the shortest paths through the network of short links. Using the resulting distance matrix, the odorants were embedded in a 3D space (Fig. 7A, the molecules are colored according to their semantic percepts). This embedding reflects the representation of the molecules in semantic space without the use of neural networks.

We then repeated this procedure for the 96 DeepNose features derived from the same odorants (Fig. 7B). We rotated the DeepNose odor space in 3D to align with the semantic space using the Procrustes algorithm [46], a three-parameter transformation. We observed that nearby molecules in the DeepNose odor space tended to share semantic descriptors (Fig. 3B), mirroring the semantic space. The layout of the descriptors was also similar to the odor space derived solely from semantic descriptors, with the primary axis (the first principal component, Leffingwell 1, Fig. 7) representing the odor pleasantness [38, 39]. These findings suggest that the DeepNose network, through training, constructs a semantic space for odorants that could be discoverable without a neural network. In addition, the DeepNose maps molecular shapes onto semantic space, enabling it to accurately predict human olfactory percepts.

### III. CONCLUSIONS

In this study, we applied neural networks to encode the 3D molecular conformations. Our main hypothesis was that the ORs function as 3D filters activated by specific features of molecular conformations. These features can be extracted using deep neural networks trained with machine learning techniques.

Our feedforward neural network consisted of two main components. First, molecular shapes were processed by an 8-layer CNN, designed to be equivariant to discrete translations and rotations. As with the CNNs being equivariant to discrete translations, since the number of discrete rotations sampled by our network was large (n=640), our network is expected to be approximately equivariant to the continuous group or rotations as well. Next, the CNN filter responses were averaged over the space and orientations in a consolidation layer to derive what we call the DeepNose features. These features were then passed through a three-layer fully connected network (MLP). The entire network was trained simultaneously on four datasets to predict binary human olfactory percepts.

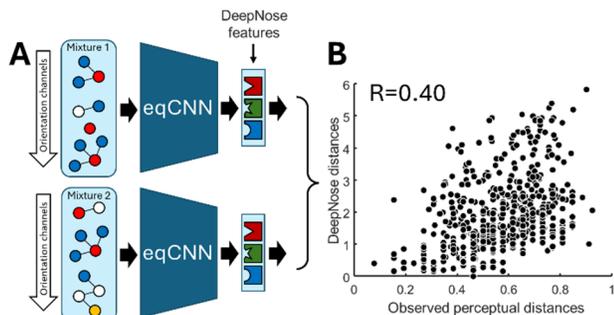

Fig. 8. Applicattion of the DeepNose to mixtures. (A) Each monomolecular component can be sent through the eqCNN via an independent orientation channel (dimension 5, Fig. 2). (B) After the DeepNose features are computed for each mixture in the pair, their Euclidean distance predicted pairwise perceptual mixture distance observed in Refs. [32-34].

The DeepNose features have demonstrated the properties that justify their comparison to the average responses of the ORs to odorants, as represented by the OSN activities. First, the values of the DeepNose features were invariant, by design, to a molecule's discrete rotations. Similarly, the OSN activities, which represent the responses of the same OR type to odorants, are not expected to be sensitive to the orientations or positions of odorant molecules. Second, in the trained model, the DeepNose features were predictive of the olfactory percepts, similarly to how the human olfactory system derives odor-evoked percepts based on the OSN/OR activities. Finally, the DeepNose features were inferred from the interactions between molecules and convolutional features of increasing complexity. This may reflect the activation of substructures within OR proteins, ranging from the displacements of individual amino acids to assemblies of amino acids and secondary structures.

Our network has demonstrated competitive performance compared to the best GNN-based models (Table 1, Fig. 3B). The performance of state-of-the-art (SOTA) networks was statistically indistinguishable from that of DeepNose, with leading models achieving an AUROC value of approximately 0.9. Unlike GNNs, our model naturally distinguishes between different 3D conformations of the same molecule (stereoisomers) and can seamlessly handle mixtures. The zero-shot performance of DeepNose on mixture data has shown a Pearson correlation of R=0.40 between predicted and measured perceptual distances. Our network's performance could significantly improve with direct training on mixture data. These strengths arise from designing our network to align closely with biological systems which interact with molecular shapes and can distinguish stereoisomers. Moreover, DeepNose could be trained on odor responses from various components of the olfactory system, such as ORs or cortical neurons, as such data becomes available.

In conclusion, we developed DeepNose, a CNN equivariant to discrete translations and rotations, capable of interacting with odor molecules in 3D and predicting human olfactory percepts. The network can distinguish between stereoisomers and respond to odorant mixtures, reflecting the behavior of various components of the biological olfactory system.

## IV. Acknowledgments

We thank Joel Mainland, Alexander Wiltschko, Richard Gerkin, and Terry Acree for making the datasets available. We thank Ngoc Tran, Daniel Kepple, Benjamin Cowley, and Dmitry Rinberg for numerous helpful discussions. This study was supported by the NIH BRAIN Initiative grant U19NS112953 and the Swartz Foundation for Computational Neuroscience.

## V. Methods

*Datasets.* We trained our model using four perceptual datasets that link individual molecules to semantic descriptors ("words") that describe their smells. These datasets included the Leffingwell dataset (3366 molecules, 113 descriptors), the GoodScents dataset (4205 molecules, 377 descriptors), the Arctander dataset (2814 molecules, 90 descriptors), and the Flavornet dataset (716 molecules, 74 descriptors), available through the Pyrfume repository [47]. In the Arctander dataset, we manually filled the missing semantic descriptors and compound (molecule) identifiers (CIDs) using Arctander's original work [19]. The union of these datasets contained 6821 unique molecules that we have split into five folds for cross-validation. To make this split, we used the iterative stratification algorithm [48] that preserved the second-order statistics of the labels (semantic descriptors) and ensured that the same molecules did not appear in different folds.

*Preprocessing.* To generate the inputs for our model, we obtained the 3D coordinates of atoms in individual molecules based on their CID identifiers using PubChemPy, a Python interface for the PubChem chemical information database. For each molecule, we generated 640 orientations by rotating the molecule around the center of the coordinates. To define the rotations that evenly cover all possible directions, we chose 64 points evenly spread on the unit sphere (a precomputed Thomson's problem solution [49]), computed the rotations that move a unit vector (0, 0, 1) to each of these points (using Kabsch algorithm via scipy.spatial.transform.Rotation class), and combined them with ten rotations of 360 / 10 = 36 degrees around the resulting direction, leading to the total of $64 \times 10 = 640$ rotation matrices. To generate raster images of all molecules, we defined a 5D array. The first (orientation) dimension corresponded to 640 orientations of each molecule;

the second (atom) dimension defined six types of atoms considered in our model (C, H, O, N, S, Cl); the remaining three dimensions contained a 3D raster image covering the space of 18×18×18Å with the step size of 1Å; each raster image has represented the positions of all atoms of a single type (e.g. carbon, C) belonging to a certain molecule at a certain orientation. We defined the coordinates of these atoms in raster images by rounding their actual coordinates, after rotation, to the nearest integer value.

*Model.* We passed these inputs through the model. In doing so, we ensured that different orientations did not interact, i.e. representations of different orientations of the same molecule were passed as parallel streams. This implied that there was complete weight sharing between orientations. We first passed the inputs through a CNN consisting of 4 convolutional blocks. Each block consisted of 2 convolutional layers, each followed by a 3D batch normalization layer and a ReLU nonlinearity. Each convolutional layer has convolutional filters with kernels of size $3\times3\times3$. The number of the convolutional filters was fixed within each convolutional block and was equal to 12, 24, 48, and 96 respectively. Between convolutional blocks, we used maximum pooling layers of size $2\times2\times2$. After the stack of convolutional layers, we used an average pooling layer of size $2\times2\times2$ that produced one-dimensional "embedding" arrays (with 96 entries) for each molecule at each orientation. We then decomposed the "batch" and the "orientation" dimensions and averaged the embeddings over 640 orientations, resulting in a one-dimensional "embedding" array (with 96 entries) for each molecule, encompassing all orientations. The 96-entry embedding (which we call the DeepNose features vector) was then sent to a stack of fully connected layers. It started with a dropout layer (20% of features, randomly selected each time, were occluded during the training iterations) followed by a hidden layer with 256 units (ReLU activation) and an output layer of 654 units (linear activation).

*Training.* The outputs of the model were then compared to concatenated vectors of semantic descriptors ("words") that described respective molecules in all four datasets (113 + 377 + 90 + 74 = 654 descriptors). Dataset labels were presented as binary vectors describing whether each of the words in the dataset vocabulary applied to a given molecule's smell. If the molecule was not present in a dataset, a binary vector was replaced with a vector of zeros that was then ignored during the learning phase. We backpropagated the difference between the model's outputs and dataset labels, masking the backpropagation to the datasets where the information about the input molecule was available. To train the model, we used the binary cross entropy loss with logits (masked to present datasets) using the ADAM optimizer with the learning rate of 3e-4 over 100 epochs on an Nvidia Tesla V100s GPU using PyTorch 2.0.0. We used 5 cross-validation splits where, for each of the four datasets, we computed the average AUROC over the semantic descriptors within the dataset.

*Ensemble averaging.* To address potential bias in a single run of the model due to the training procedure, for each cross-validation split, we trained 5 networks with different random weight initializations and data shuffling. The average output of the 5 networks on the same test set was taken as output for the ensemble.

*Contributions of molecular features to individual descriptors.* To estimate each atom's contribution to semantic percepts, we simulated atom occlusions by removing atoms, one at a time, from each molecule and passing the occluded molecules through the DeepNose. Comparing the predictions for the original and occluded molecules described each atom's impact on each of the perceptual descriptors. To observe such atomic contributions across scales, we scaled these differences and applied a logistic function to constrain the output to the range [-1, 1].

*Classification.* To establish the thresholds for binarizing the predicted semantic descriptors, we sorted the DeepNose outputs for the training data and identified the values that equalized true and false positive rates in the binarized data for every percept independently in the training data. These threshold values were then applied to classify the test data.

*Application to mixtures.* To test the zero-shot generalizability of our model to odor mixture data, we used four datasets describing perceptual distances between pairs of odor mixtures. These datasets included the Bushdid [32], Snitz [33] (two datasets), and Ravia [34] datasets, available as a combined dataset through the DREAM Olfactory Mixtures Prediction Challenge [50]. We used the training split of this data, describing 500 perceptual distances between 703 mixtures of monomolecular odorants (1-43 single molecules per mixture). To form the predictions for the perceptual distances between odor mixtures, we computed DeepNose features for mixtures as follows. For constituent molecules of each mixture, we computed the DeepNose features (defined as the outputs of the CNN part of our model; 1-43 feature vectors per mixture) using a pretrained model. We averaged the DeepNose features of individual molecules to form the embeddings for the mixtures (703 embeddings with 96 entries each). This procedure was functionally equivalent to randomly sampling constituent molecules of a mixture to form the inputs along the "orientation" dimension of a single pretrained DeepNose model and using its output (with 96 entries) as a joint DeepNose feature for the mixture. We then computed pairwise Euclidean distances between 500 pairs of these DeepNose feature vectors of the mixtures and compared them to the perceptual distances reported in the datasets by computing the Pearson correlation.